\documentclass{article}
\usepackage[nonatbib,final]{nips_2018}
\usepackage[utf8]{inputenc} 
\usepackage[T1]{fontenc}    
\usepackage{hyperref}       
\usepackage{url}            
\usepackage{booktabs}       
\usepackage{amsfonts}       
\usepackage{nicefrac}       
\usepackage{microtype}      
\usepackage{multirow}
\usepackage{subfig}
\usepackage{amsmath,graphicx}

\title{Deep Within-Class Covariance Analysis for\\
Robust Audio Representation Learning}

\author{
  Hamid Eghbal-zadeh$^{1,2}$
  \ \ \ \ \ \ \ \ \ \ \ 
  Matthias Dorfer$^2$
  \ \ \ \ \ \ \ \ \ \ \ 
  Gerhard Widmer$^{1,2}$  \\ \\
  $^1$~LIT AI Lab \& $^2$~Institute of Computational Perception\\
  Johannes Kepler University of Linz, Austria\\
  \texttt{\{hamid.eghbal-zadeh, matthias.dorfer, gerhard.widmer\}@jku.at}
}

\begin{document}

\maketitle

\begin{abstract}
Deep Neural Networks (DNNs) are known for excellent performance in supervised tasks such as classification. 
Convolutional Neural Networks (CNNs), in particular, can learn effective features and build high-level representations that can be used for classification, but also for querying and nearest neighbor search.
However, CNNs have also been shown to suffer from a performance drop when the distribution of the data changes from training to test data.
In this paper we analyze the internal representations of CNNs and observe that the representations of unseen data in each class, spread more (with higher variance) in the embedding space of the CNN compared to representations of the training data.
More importantly, this difference is more extreme if the unseen data comes from a shifted distribution.
Based on this observation, we objectively evaluate the degree of representation's variance in each class by applying eigenvalue decomposition on the within-class covariance of the internal representations of CNNs and observe the same behaviour. 
This can be problematic as larger variances might lead to mis-classification if the sample crosses the decision boundary of its class.
We apply nearest neighbor classification on the representations and empirically show that the embeddings with the high variance actually have significantly worse KNN classification performances, although this could not be foreseen from their end-to-end classification results.
To tackle this problem, we propose \emph{Deep Within-Class Covariance Analysis (DWCCA)}, a deep neural network layer that significantly reduces the within-class covariance of a DNN's representation, improving performance on unseen test data from a shifted distribution.
We empirically evaluate DWCCA on two datasets for Acoustic Scene Classification (DCASE2016 and DCASE2017).
We demonstrate that not only does DWCCA significantly improve the network's internal representation, it also increases the end-to-end classification accuracy, especially when the test set exhibits a slight distribution shift.
By adding DWCCA to a VGG neural network, we achieve around 6 percentage points improvement in the case of a distribution mismatch.
\end{abstract}

\section{Introduction}
\label{sec:intro}
Convolutional Neural Networks (CNNs) are the state of the art in many supervised learning tasks such as classification, and using the power of convolutional layers, CNNs can learn useful features that are often superior to engineered features, and build internal representations that can achieve high classification performance.

It has been shown that CNNs have a surprising ability to fit data, so much so that they can even perfectly learn from data with random labels~\cite{zhang2016understanding}.
But of course, memorising the training data is not sufficient: a model is expected to generalize to unseen data points.
Additionally, a robust model has to be able to not only deal with unseen data points that are similar to the training set, but also cope with unseen data points that may come from a slightly different distribution than the training data (\emph{distribution mismatch}).
When there is a distribution shift between the training and test sets, robustness of the model's representation becomes more important as it has to classify or embed data points that are quite different from the ones it has observed in the training set.

In this paper, we investigate this by using a well-known DNN architecture (VGG~\cite{simonyan2014very}) that is adapted for audio classification~\cite{eghbal2016cp} and is widely used among researchers.
We evaluate VGG on data with as well as without distribution mismatch and observe that while VGG exhibits a reasonable performance on the data without distribution mismatch, its performance significantly drops when tested on data from a shifted distribution.

We start by analyzing the internal representations of the network by using visualisations.
As will be seen in the first (a-c) and the 3rd rows (g-i) of Figure~\ref{fig:embeddings}, the network's internal representations in each class spread more in the embedding space for the unseen data (validation or test) compared to the training data. 
This is even more extreme when the unseen data comes from a shifted distribution (i).

For an objective evaluation of the amount of the representation's variance in each class, we compute the within-class covariance of the representations of the network for each class, and we apply eigenvalue decomposition to compute the eigenvalues of each class's covariance matrix.
We then report the sorted eigenvalues of the within-class covariance of the representations in Figure~\ref{fig:wcc}.
As the blue curves show, the eigenvalues in unseen data of validation (b and e) and test (c and d) have considerably higher ranges than train data (a and d) for all the datasets we used.

To better understand the effect of such high variance in the quality of generalisation in the representations of our network, we carry out K-nearest neighbor (KNN) experiments on the dataset without, and the dataset with distribution shift.
As the results in Figure~\ref{fig:knn} show, the performance degredation from validation (c) compared to test (d) in case of distribution mismatch is significantly higher compared to the performance drop from validation (a) to test (b) when the test data comes from a similar distribution.
This observation is also aligned with what we observed in the visualisations from Figure~\ref{fig:embeddings} that showed the data is more spread than validation data, when coming from a shifted distribution.

To tackle this problem, we propose \emph{Deep Within-Class Covariance Analysis (DWCCA)}, a deep neural network layer that reformulates the conventional Within-Class Covariance Normalization (WCCN)~\cite{generalized_Hatch06} as a DNN-compatible version. 
DWCCA is trained end-to-end using back-propagation, can be placed in any arbitrary position in a DNN, and is capable of significantly reducing the within-class covariance of the internal representation in a DNN.

We empirically show that DWCCA significantly reduces the within-class covariance of the DNN's representations, in both cases.
Further, we evaluate the generalization quality of the DNN's representations after applying DWCCA by performing nearest neighbor classification on its representations.
Our results show that DWCCA significantly improves the nearest neighbor classification results in both cases, hence improving the generalization quality of the representations.
And finally we report the end-to-end classification results of the trained models on an acoustic scene classification task, using data from the annual IEEE Challenges on Detection and Classification of Acoustic Scenes and Events (DCASE). It turns out that the classification results for the dataset with distribution shift are significantly improved by integrating the DWCCA layer, while the performance on the dataset without distribution mismatch stayed the same.

\section{Related Work}

The characteristics of the representations learned in CNNs can be influenced by many factors such as network architecture and the training data. The authors in~\cite{kornblith2018better} investigated how architecture topologies affect the robustness and generalization of a representation and showed which representations from different architectures better transfer to other datasets and tasks. 

While the topology of the architecture influences the generality of its representations,
several authors proposed methods that can improve the internal representation of a DNN.
\cite{Dorfer2015} proposed a loss which learns linearly separable latent representations on top of a DNN, by maximising their smallest Linear Discriminant Analysis (LDA) \cite{fisher1936use} eigenvalues.
And in~\cite{Andrew2013}, the authors propose creating a maximally correlated representation from different modalities.

It has been shown that stand-alone CNNs are not very successful at generalizing to unseen data points from shifted distributions in tasks such as Acoustic Scene Classification (ASC), where such distributional shifts are common. ASC is defined as classifying environments from the sounds they produce~\cite{barchiesi2014acoustic}, and often these environments (e.g, home, park, city center) may sound very different in different cities or during different times of the year.
Although in~\cite{Marchi2016,valenti2016dcase,eghbal2016cp} CNNs have shown promising results in ASC when the unseen test data has a similar distribution to the training data, in~\cite{lehner2017classifying,Zhao2017} similar CNNs that previously performed successfully in a matched distribution case, suffered significantly from the lack of generalization to a shifted distribution in the test set.

To cope with this drawback of CNNs,~\cite{Mun2017} investigated CNN manifold data augmentation using Generative Adversarial Networks~\cite{goodfellow2014generative}, while in~\cite{Hyder2017,Weiping2017,Mun2017} the authors used an ensemble of CNNs as feature extractors and processed the deep features via Support Vector Machines (SVMs), followed by a late fusion of various models. They showed that although CNNs do not generalize well in the distribution shift case, they can learn useful features that can be incorporated for building new classifiers with better generalization properties.

In this paper, we try to better understand these problems. We focus our efforts on investigating the reasons for the performance drop in CNNs when the data distribution is shifted. To this end, we analyze the internal representations in CNNs and propose DWCCA to tackle this issue.

\section{Deep Within-Class Covariance Analysis}
\label{sec:methods}
We start by introducing a common notation which will be used throughout the paper.
We then first describe Within-Class Covariance Normalization (WCCN) --a classic machine learning method--, and then show how to cast it into a deep learning compatible version. 

\subsection{Conventional Within-Class Covariance Normalization}
\label{subsec:wccn}
Let $\mathbf{W} 
$
denote a set of $N$ $d$-dimensional observations (feature vectors)
belonging to $C$ different classes $c \in \{1, ..., C\}$.
The observations are in the present case either hand crafted features (e.g. i-vectors~\cite{dehak2011front}) or any intermediate hidden representation of a deep neural network.

WCCN is a linear projection that provides an effective compensation for the within-class variability and has proven to be effectively used in combination with different scoring functions~\cite{Dehak2010,li2013speaker}.
The WCCN projection scales the feature space in the opposite direction of its within-class covariance matrix,
which has the advantage that finding decision boundaries on the WCCN-projected data becomes easier~\cite{wccn_Hatch06}.
The within-class covariance $\mathbf{S}_w$ is estimated as:
\begin{equation}
\label{eq:wccn}
\mathbf{S}_w=\frac{1}{C}\sum_{c=1}^{C} \frac{1}{N_c} \sum_{i=1}^{N_c} ({\mathbf{W}^c}_i - \overline{\mathbf{W}^c})({\mathbf{W}^c}_i - \overline{\mathbf{W}^c})^T
\end{equation}
where $\overline {\mathbf{W}^c}= \frac{1}{N_c} \sum_{i=1}^{N_c} {\mathbf{W}^c}_i$ is the mean feature vector of class $c$
and ${\mathbf{W}^c}_i$ are the samples belonging to this class.
$N_c$ is the number of observations of class $c$ in the training set.
We use the inverse of matrix $\mathbf{S}_w$ to normalize the direction of the projected feature vectors.
The WCCN projection matrix $\mathbf{B}$ can be estimated using the Cholesky decomposition as
$\mathbf{B} = \mathit{cholesky}(\mathbf{S}_w^{-1})$.
\subsection{Deep Within-Class Covariance Analysis (DWCCA)}
\label{subsec:swccn}
\begin{figure}[ht]
\centering
\includegraphics[height=0.2\textwidth]{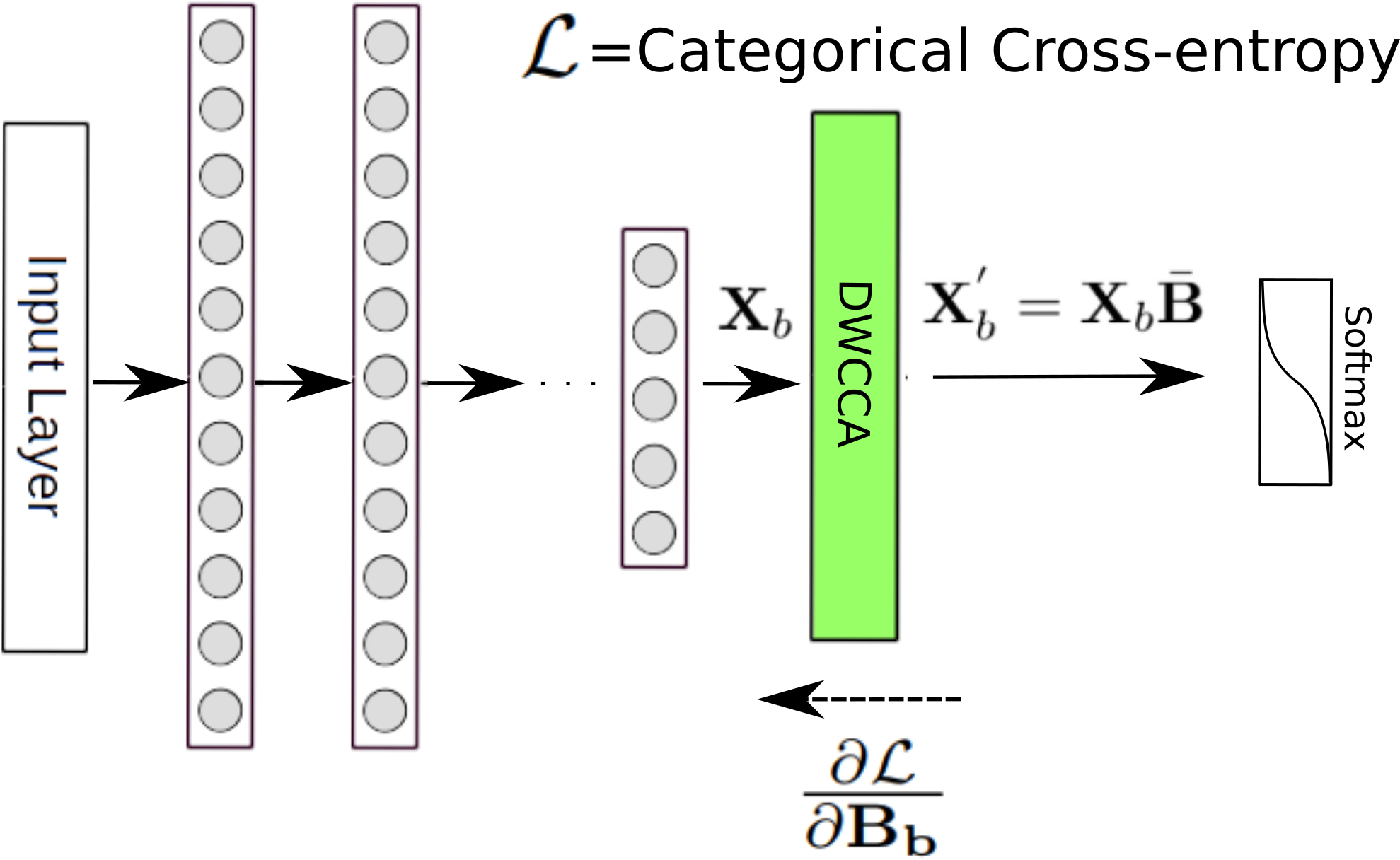}
\caption{Diagram of the proposed layer in a DNN architecture.}
\label{fig:diag}
\end{figure}

Based on the definitions above we propose Deep Within-Class Covariance Analysis (DWCCA), a DNN-compatible formulation of WCCN.
The parameters of our networks are optimized with Stochastic Gradient Descent (SGD), and trained with mini-batches.
The deterministic version of WCCN described above is usually estimated on the entire training set, which by the definition of SGD is not available in the present case.
In the following we propose a DWCCA Layer which helps to circumvent these problems.
Figure~\ref{fig:diag} shows a schematic sketch of how the DWCCA Layer can be incorporated in a neural network, and its gradient flow.

Instead of computing the within-class covariance matrix $\mathbf{S}_w$
on the entire training set $\mathbf{W}$
we provide an estimate $\hat{\mathbf{S}}_b$ on the observations $\mathbf{W}_b$ of the respective mini-batches.
Given this estimate we compute the corresponding mini-batch projection matrix $\hat{\mathbf{B}}_b$
and use it to maintain a moving average projection matrix $\bar{\mathbf{B}}$ as $\bar{\mathbf{B}} = (1-\alpha) \bar{\mathbf{B}} + \alpha \hat{\mathbf{B}}_b$.

This is done similarly when computing the mean and standard deviation for normalising the activations in batch normalization \cite{LoffeS2015BatchNorm}.
The hyper parameter $\alpha$ controls the influence of the data in the current batch $b$ on the final DWCCA projection matrix.
The output of this processing step is the DWCCA projected data $\mathbf{W}^{'}_b = \mathbf{W}_b \bar{\mathbf{B}}$ of the respective mini-batch.

The DWCCA Layer can be seen as a special dense layer with a predefined weight matrix, the projection matrix $\bar{\mathbf{B}}$, with the difference that the parameters are computed via the activations of the previous layer, and not learned via SGD.
The proposed covariance normalization is applied directly during optimization. 
For training our network with back-propagation, we need to establish gradient flow.
To this end, we implement the DWCCA Layer using the automatic differentiation framework Theano~\cite{bergstra_2010_theano}
which already provides the derivatives of matrix inverses and the Cholesky decomposition.
We refer the interested reader to~\cite{smith1995differentiation} for details on this derivative.

\section{Empirical Results}
\label{sec:exp}

In this section, we detail our experimental setup and present our empirical results.

\subsection{Dataset}
\label{subsec:db}
To evaluate the performance of CNNs in situations with and without distribution mismatch, we use the TUT Acoustic Scenes 2016 (DCASE2016)~\cite{mesaros2016tut} and TUT Acoustic Scenes 2017 (DCASE2017)~\cite{mesaros2017dcase} datasets.

The DCASE2016 dataset consists of 15 different
acoustic scenes: lakeside beach, bus, cafe/restaurant, car, city
center, forest path, grocery store, home, library, metro station,
office, urban park, residential area, train, and tram. 
All audio material was collected as 3-5 minutes length recordings and then was cut into segments of 30 seconds length.
The DCASE2017 dataset consists of the same 15 classes and uses the same recordings in its development set (training and validation) as used for DCASE2016 development set. 
But these recordings were split into only 10s long segments instead of 30s, which makes the learning and classification task harder. A new evaluation set (unseen test) was recorded for DCASE2017.
The reason for choosing these two datasets is that DCASE2016 and DCASE2017 are based on the same development (= training and validation) data, but have different evaluations (= independent test) sets. 
While the test data for DCASE2016 were collected jointly with the development data, new test recordings were collected for DCASE2017, at a later time, in different places.
Hence, there is a distribution mismatch between development and test data in the case of DCASE2017, as the many environments sound different from those in the development set (which will also show, indirectly, in our experimental results). This is detailed in the report of DCASE2017~\cite{mesaros2017dcase}.

In the following, we will use DCASE2016 to represent a data scenario without distribution shift, and DCASE2017 as the distribution shift case.
In all experiments, we use the official 4-fold cross validation splits provided with the datasets and report CV results on all folds separately, as well as the performance of our method on the unseen test set.

\subsection{Baseline Systems and Experimental Setup}
\label{subsec:bl}

As explained before, we use a VGG-Style CNN proposed in~\cite{eghbal2016cp} which uses audio spectrograms as inputs. We will refer to this model as \emph{vanilla}.

To evaluate the effect of our proposed method, we apply a DWCCA layer on the output of global average pooling, as can be seen in Table~\ref{tab:model_architecture}. We will refer to this model in our results as \emph{dwcca}.
The internal \emph{representations} we will analyze are computed by using the output of global average pooling layer in \emph{vanilla}, and the output of the DWCCA layer in \emph{dwcca}.

We also provide additional baselines for comparison:
We report the performance of the best end-to-end CNN model~\cite{Marchi2016} in the DCASE2016 challenge~\cite{mesaros2018detection}. This method uses an ensemble of various CNN models.
For the DCASE2017 experiments, we also choose the best-performing end-to-end CNN from the DCASE2017 challenge~\cite{mesaros2017dcase} as an additional baseline to vanilla:
~\cite{Zhao2017} that uses an ensemble fusion of various ResNets~\cite{he2016deep} using several channels of audio (left, right and difference).

We report classification results of our end-to-end CNNs on all folds, and on the unseen test set. For the unseen test, we average the probabilities produced by the 4 models trained on the individual training parts in CV folds from the development data.
Additionally, we report \emph{calibrated} results where we use a linear logistic regression model in each fold to regress the prediction probabilities of each model to the true labels. This method is better known as \emph{late fusion} and proved successful in~\cite{eghbal2016cp}. The \emph{probabilistic averaging} for unseen test is done similarly for late fusion, as explained above.

The setup used in \emph{all} of our experiments are as follows: The initial learning rate is set to $0.0001$, and the ADAM optimizer is used~\cite{kingma2014adam}.
We choose a similar $\alpha$ (0.1) to the one used in batchnorm. 
A model selection with max. patience of 20 is applied and the learning rate is halved in each step of maximum patience.
The architecture of our models is specified in Table~\ref{tab:model_architecture}.
All models are trained with stratified batch iterators and batch size of 75. In all our experiments, the spectrogram of the whole audio segment (30 sec in DCASE2016 and 10 sec in DCASE2017) was fed to our models.
All spectrograms in our experiments are extracted with the \emph{madmom} library~\cite{bock2016madmom} using the mono signal, and the \emph{Lasagne} library~\cite{dieleman2015lasagne} was used for building neural network architectures.
The DWCCA Layer was implemented as a Lasagne-compatible layer in Theano~\cite{bergstra2011theano}.

\begin{table}[t!]
\caption{Model Specifications. BN: Batch Normalization, ReLu: Rectified Linear Activation Function, CCE: Categorical Cross Entropy.}
\label{tab:model_architecture}
\begin{center}
\begin{tabular}{c}
Input $1 \times 938$ (DCASE2016) or $313$ (DCASE2017) $\times 149$ \\
\hline
$5\times5$ Conv(pad-2, stride-2)-$32$-BN-ReLu \\
$3\times3$ Conv(pad-1, stride-1)-$32$-BN-ReLu \\
$2\times2$ Max-Pooling + Drop-Out($0.3$) \\
\hline
$3\times3$ Conv(pad-1, stride-1)-$64$-BN-ReLu \\
$3\times3$ Conv(pad-1, stride-1)-$64$-BN-ReLu \\
$2\times2$ Max-Pooling + Drop-Out($0.3$) \\
\hline
$3\times3$ Conv(pad-1, stride-1)-$128$-BN-ReLu \\
$3\times3$ Conv(pad-1, stride-1)-$128$-BN-ReLu \\
$3\times3$ Conv(pad-1, stride-1)-$128$-BN-ReLu \\
$3\times3$ Conv(pad-1, stride-1)-$128$-BN-ReLu \\
$2\times2$ Max-Pooling + Drop-Out($0.3$) \\
\hline
$3\times3$ Conv(pad-0, stride-1)-$512$-BN-ReLu \\
Drop-Out($0.5$) \\
$1\times1$ Conv(pad-0, stride-1)-$512$-BN-ReLu \\
Drop-Out($0.5$) \\
\hline
$1\times1$ Conv(pad-0, stride-1)-$15$-BN-ReLu \\
Global-Average-Pooling \\
\hline
DWCCA (if applied) \\
\hline
$15$-way Soft-Max
\end{tabular}
\end{center}
\end{table}

\subsection{Experiments Performed}
\label{subsec:experiments}

We carried out various experiments to demonstrate the issue we encounter in ASC with CNNs. 
First, we train our models on DCASE2016 and DCASE2017.
For a better understanding of the issues in the representation learned by the CNN, we provide a \emph{visualization} of the VGG representations after being projected into 2D via Principal Component Analysis (PCA)~\cite{jolliffe2011principal}. The result can be found in Figure~\ref{fig:embeddings}.

For analysing the \emph{within-class covariance} of the representations, we apply an \emph{eigenvalue decomposition} on the covariance matrix of the representations from each class. This technique is also used in other studies such as~\cite{le1991eigenvalues} to investigate the dynamics of learning in neural networks.
These eigenvalues are a good indicator of the covariance of the representations in each class. If high, it means that the representations on that specific class have a high covariance, and if the difference between the highest eigenvalue and the lowest eigenvalue is high it also indicates that the variance of the representation in that class is not spread equally in all dimensions.
These results can be found in Figure~\ref{fig:wcc}. The shade in this plot represents the variance of eigenvalues over different classes.

To indirectly assess the structure of the internal representation spaces learned, we carry out a \emph{k-nearest-neighbor classification experiment} in these representation spaces; the K-nearest neighbor results on both datasets will be shown in Figure~\ref{fig:knn}.

Finally, we report the \emph{end-to-end classification results} on both datasets, for various methods; these can be found in Table~\ref{tab:full_results}.
For a deeper understanding of the proposed method, we additionally provided the class-wise f-measure of DWCCA and the baseline VGG (vanilla) in Table~\ref{tab:class_wise}.

\subsection{Results and Discussions}
\label{subsec:results}

\begin{figure}[t!]
\centering
\subfloat[\parbox{.6in}{DCASE2016 training (vanilla)}]{\label{fig:embeddings_16_train_vanil}{\includegraphics[height=0.12\textwidth]{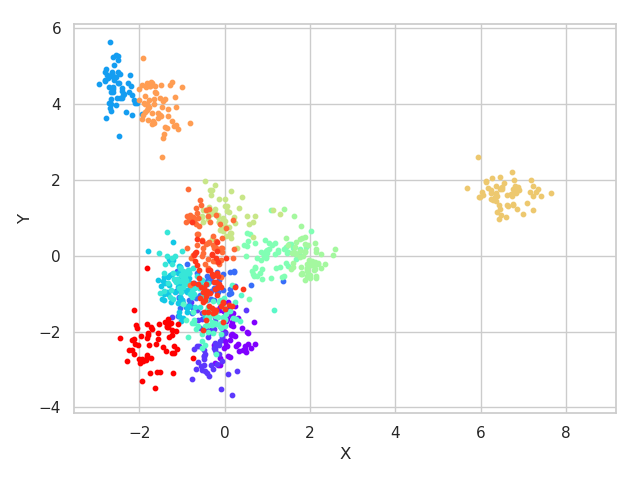} }}%
\subfloat[\parbox{.6in}{DCASE2016 validation (vanilla)}]{\label{fig:embeddings_16_val_vanil}{\includegraphics[height=0.12\textwidth]{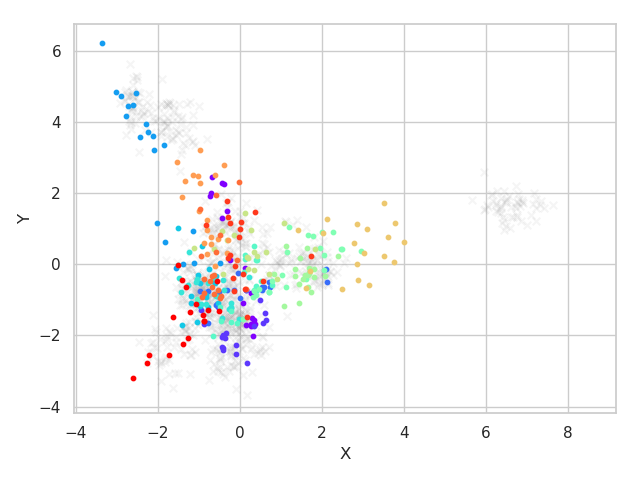} }}%
\subfloat[\parbox{.6in}{DCASE2016 unseen test (vanilla)}]{\label{fig:embeddings_16_test_vanil}{\includegraphics[height=0.12\textwidth]{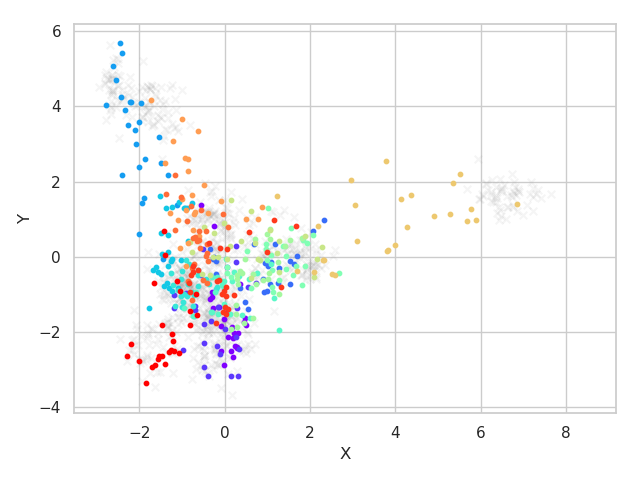} }}%
\subfloat[\parbox{.6in}{DCASE2016 training (dwcca)}]{\label{fig:embeddings_16_train_dwc}{\includegraphics[height=0.12\textwidth]{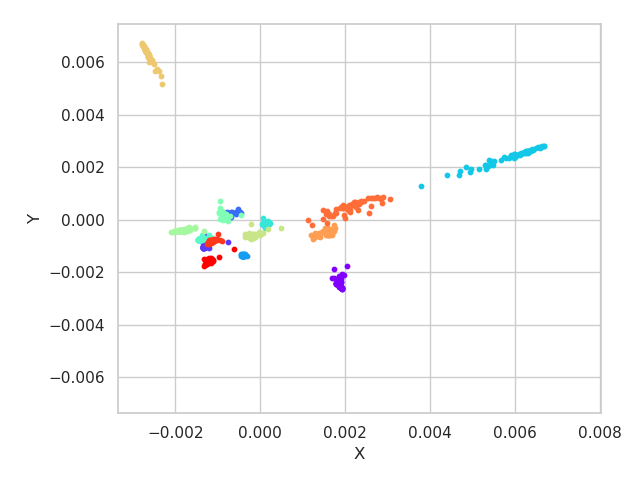} }}%
\subfloat[\parbox{.6in}{DCASE2016 validation (dwcca)}]{\label{fig:embeddings_16_val_dwc}{\includegraphics[height=0.12\textwidth]{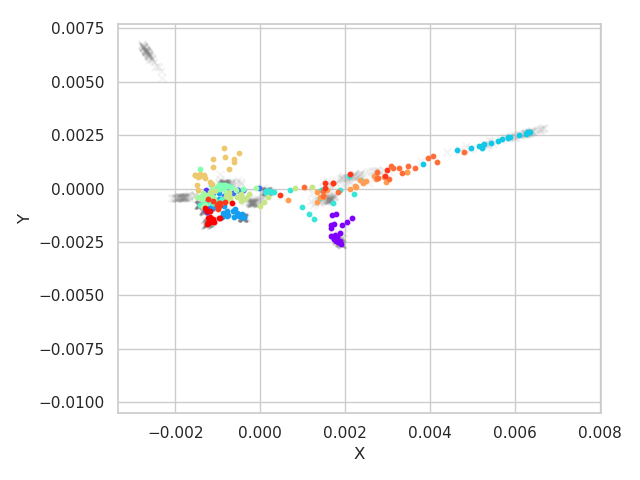} }}%
\subfloat[\parbox{.6in}{DCASE2016 unseen test (dwcca)}]{\label{fig:embeddings_16_test_dwc}{\includegraphics[height=0.12\textwidth]{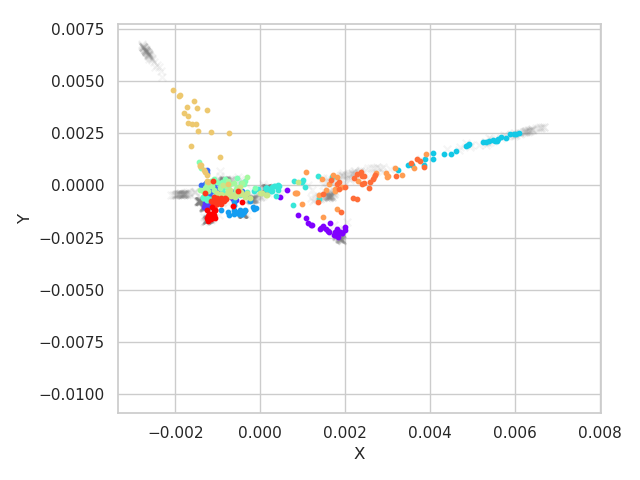} }}%
\\
\subfloat[\parbox{.6in}{DCASE2017 training (vanilla)}]{\label{fig:embeddings_17_train_vanil}{\includegraphics[height=0.12\textwidth]{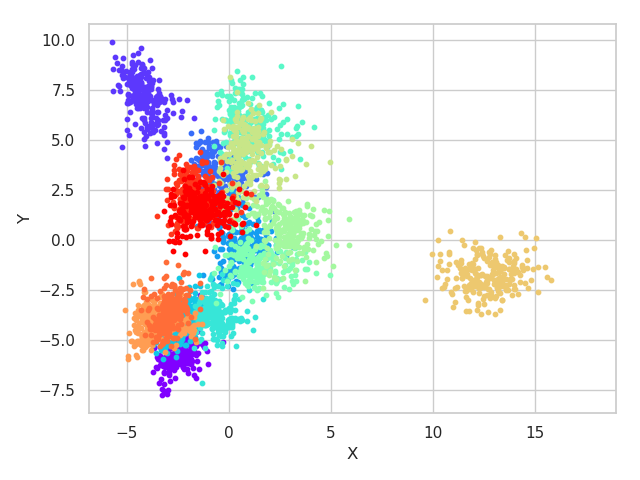} }}%
\subfloat[\parbox{.6in}{DCASE2017 validation (vanilla)}]{\label{fig:embeddings_17_val_vanil}{\includegraphics[height=0.12\textwidth]{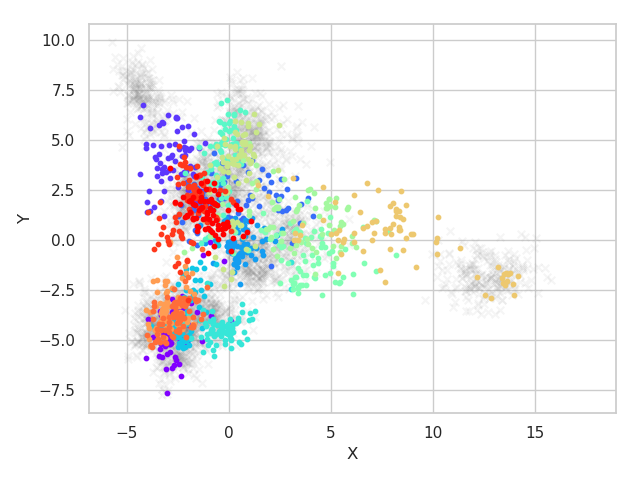} }}%
\subfloat[\parbox{.6in}{DCASE2017 unseen test (vanilla)}]{\label{fig:embeddings_17_test_vanil}{\includegraphics[height=0.12\textwidth]{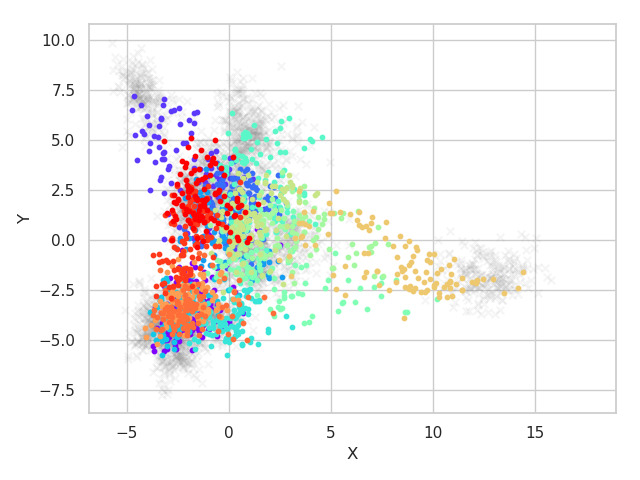} }}%
\subfloat[\parbox{.6in}{DCASE2017 training (dwcca)}]{\label{fig:embeddings_17_train_dwc}{\includegraphics[height=0.12\textwidth]{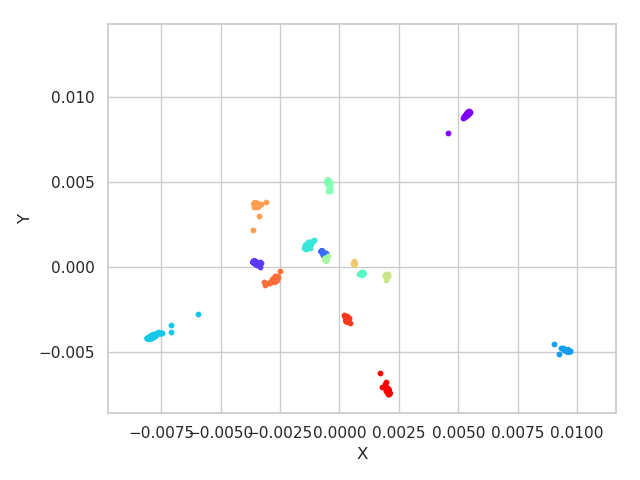} }}%
\subfloat[\parbox{.6in}{DCASE2017 validation (dwcca)}]{\label{fig:embeddings_17_val_dwc}{\includegraphics[height=0.12\textwidth]{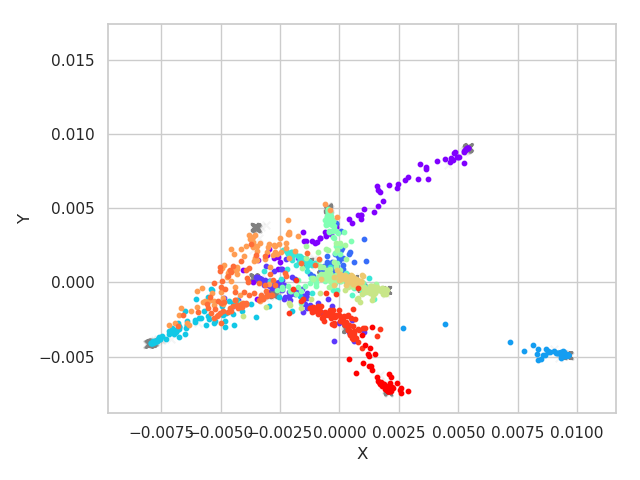} }}%
\subfloat[\parbox{.6in}{DCASE2017 unseen test (dwcca)}]{\label{fig:embeddings_17_test_dwc}{\includegraphics[height=0.12\textwidth]{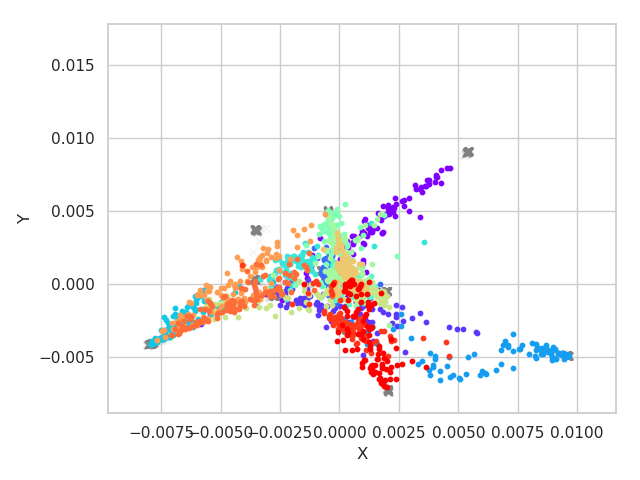} }}%
\caption{DNN representations projected to 2D by PCA. 
The gray embeddings in low opacity (available only in validation and test plots) represent the training data. Note the distribution mismatch for vanilla.}
\label{fig:embeddings}
\end{figure}

In Figure~\ref{fig:embeddings}, the network's internal representations in each class are projected into 2D via PCA and each class is represented by a different colour.
Looking at first (a-c) and second (d-f) row, it can be seen that for the dataset without mismatched distribution the embeddings of unseen data (validation and test) are spread less after applying DWCCA.
Also comparing the unseen embeddings to the training embeddings (with lower opacity and in grey) it can be seen that the unseen embeddings projected closer to the training embeddings after applying DWCCA.
Comparing third (g-i) and fourth (j-l) row, it can be seen that for the case of a distribution shift DWCCA also reduces the variance of the embeddings in each class, resulting in them being embedded closer to the training embeddings (grey).
This suggests that this property can improve the generalisation of the representations. We will empirically evaluate this hypothesis later in this section by applying KNN classification on the representations.

\begin{figure*}[t!]
\centering
\subfloat[\parbox{.7in}{DCASE2016 training set}]{\label{fig:wcc_16_train}{\includegraphics[height=0.12\textwidth]{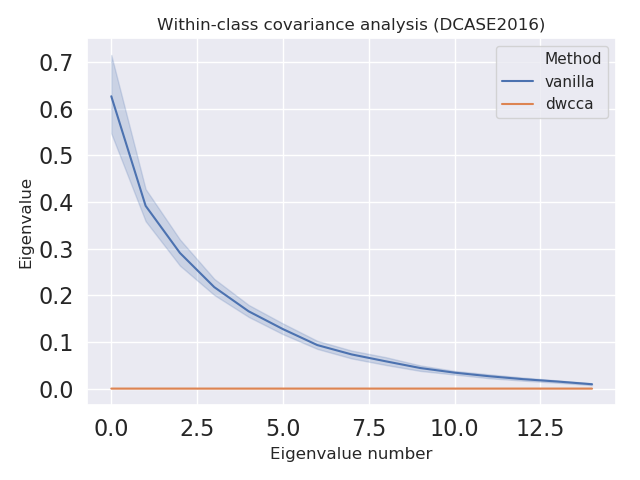} }}%
\subfloat[\parbox{.7in}{DCASE2016 validation set}]{\label{fig:wcc_16_val}{\includegraphics[height=0.12\textwidth]{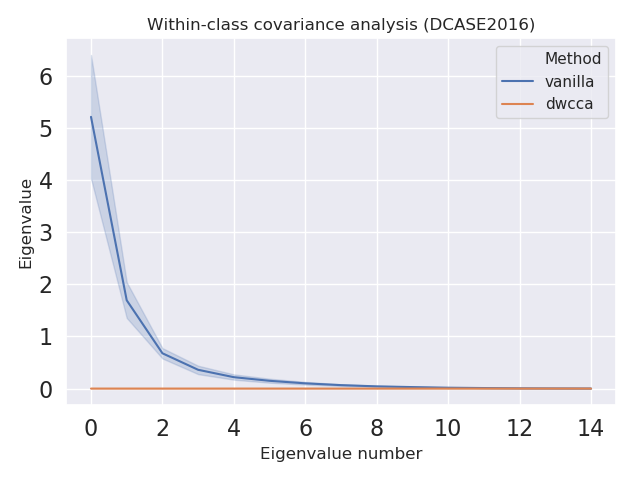} }}%
\subfloat[\parbox{.7in}{DCASE2016 unseen test set}]{\label{fig:wcc_16_test}{\includegraphics[height=0.12\textwidth]{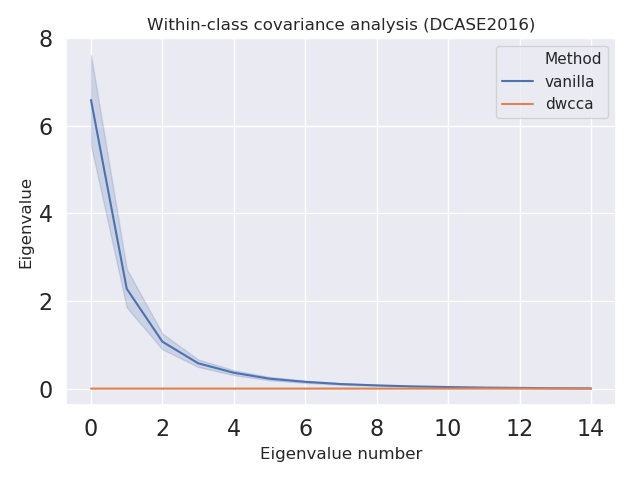} }}%
\subfloat[\parbox{.7in}{DCASE2017 training set}]{\label{fig:wcc_17_train}{\includegraphics[height=0.12\textwidth]{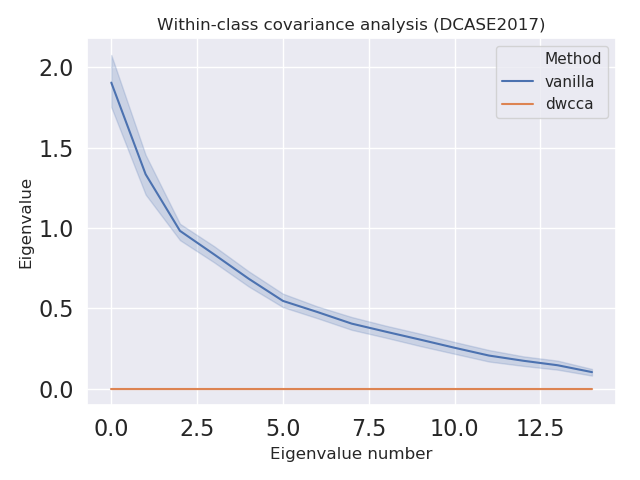} }}%
\subfloat[\parbox{.7in}{DCASE2017 validation set}]{\label{fig:wcc_17_val}{\includegraphics[height=0.12\textwidth]{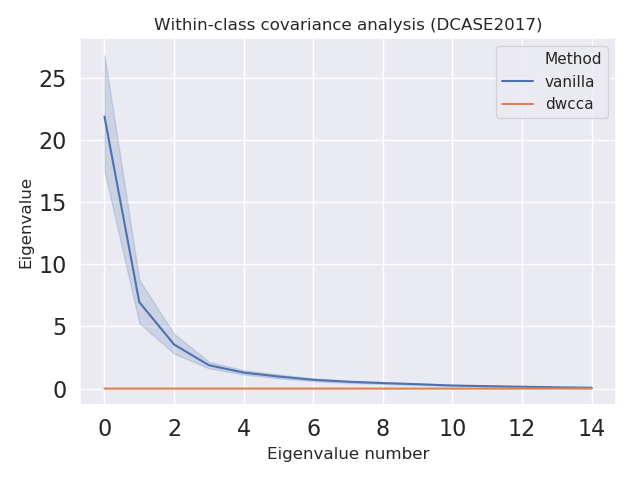} }}%
\subfloat[\parbox{.7in}{DCASE2017 unseen test set}]{\label{fig:wcc_17_test}{\includegraphics[height=0.12\textwidth]{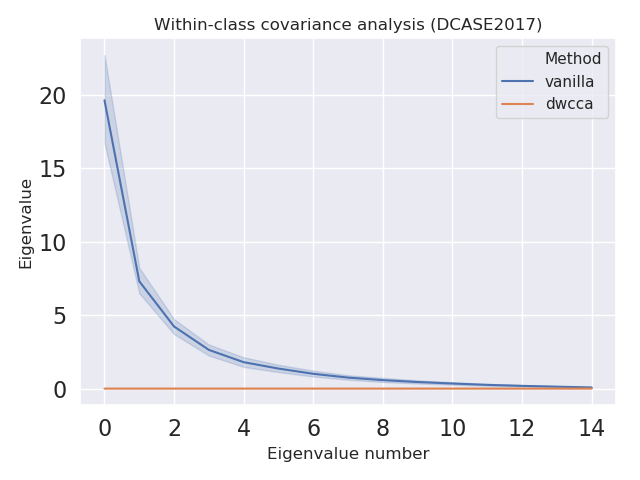} }}%
\caption{Eigenvalues of the covariance of the network representation.
Shade represents the variance over different classes.}
\label{fig:wcc}
\end{figure*}

Looking at Figure~\ref{fig:wcc}, we can see that in all plots from dataset with, and dataset without distribution shift, DWCCA significantly reduces the within-class variability. This can be observed by looking at the eigenvalues of the covariances of the representations. 
An interesting observation is 
the range of eigenvalues in vanilla: In both datasets, eigenvalues have significantly larger range on unseen data (validation and test) compared to the training data.
The maximum eigenvalue in DCASE2016 is around 0.7, while the maximum eigenvalue for unseen is around 7, about 10 times more. Also the maximum eigenvalue of the train set of DCASE2017 is around 2, while the max. eigenvalue on unseen data is around 20 (10 times larger).

\begin{figure*}[t!]
\centering
\subfloat[\parbox{.7in}{\scriptsize DCASE2016 KNN results on validation set}]{\label{fig:knn_16_val}{\includegraphics[height=0.18\textwidth]{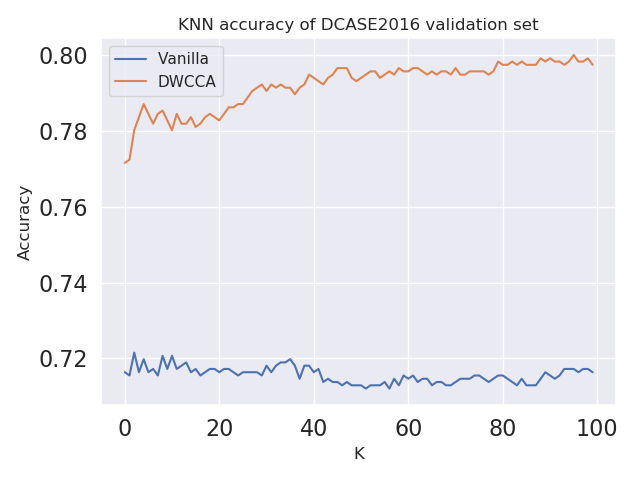} }}%
\subfloat[\parbox{.7in}{\scriptsize DCASE2016 KNN results on unseen test set}]{\label{fig:knn_16_test}{\includegraphics[height=0.18\textwidth]{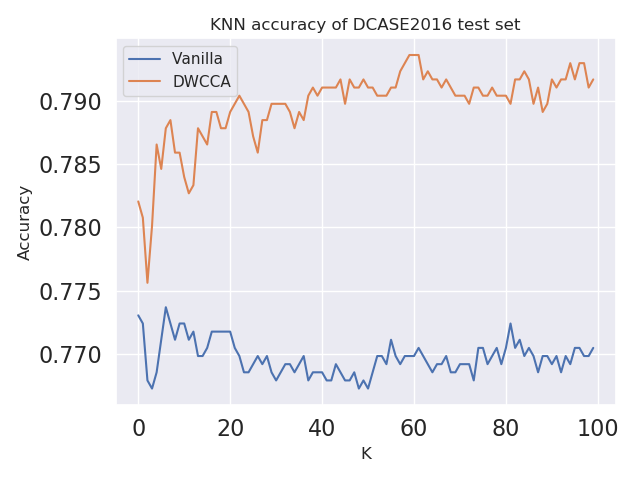} }}%
\subfloat[\parbox{.7in}{\scriptsize DCASE2017 KNN results on validation set}]{\label{fig:knn_17_val}{\includegraphics[height=0.18\textwidth]{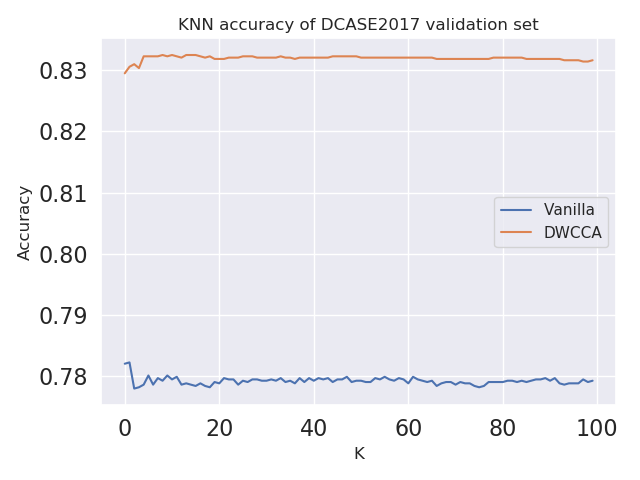} }}%
\subfloat[\parbox{.7in}{\scriptsize DCASE2017 KNN results on unseen test set}]{\label{fig:knn_17_test}{\includegraphics[height=0.18\textwidth]{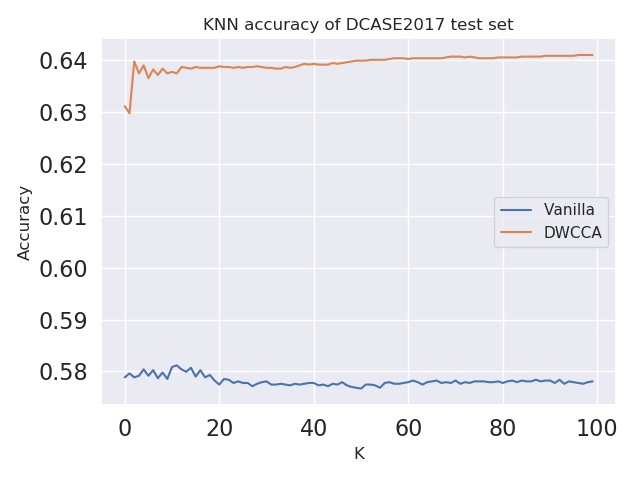} }}%
\caption{KNN results: to predict the label of each unseen data point (validation or test), its representation is assigned to the labels of its K-nearest representation from the DNN's representations of training set. Axis x represents K and y axis shows the classification accuracy for each K.}
\label{fig:knn}
\end{figure*}

By looking at the KNN results in Fig.~\ref{fig:knn} it can be seen that in both cases (mismatch / no mismatch), the KNN classification accuracy increases by adding DWCCA. Also, while the KNN performance is in a reasonable range on the validation set of both datasets, the test accuraty in the mismatch case (DCASE2017) drops significantly compared to the validation set.
Additionally it can be seen that applying DWCCA significantly improves the performance on the test set with shifted distribution, adding an improvement of about 6 percentage point, while the improvement on the test set without mismatch is around 2 percentage points.
Looking at the results of end-to-end classifications in Table~\ref{tab:full_results}, we see that the performance of vanilla on DCASE 2017 consistently and significantly improves when adding DWCCA, on all development folds as well as on the unseen test data.
We observe around 6 percentage points improvement by adding DWCCA to VGG.

Looking at the results of the dataset without mismatch, we see that although the results on all folds were improved by adding DWCCA, the results on the unseen test set do not significantly change.
This can be explained better by looking at Figure~\ref{fig:embeddings}: the embeddings of validation (b) and test (c) indicate that the test data is projected closer to the training set than the validation set. This observation suggests that the unseen test in DCASE2016 might be similar (even more similar than the validation data) to the training set. This can also be confirmed by looking at the results of the \emph{best CNN} baseline, as well as vanilla: the performances on the unseen test set are consistently higher than all the validation folds.
Hence, DWCCA could not help as there was not a large generalisation gap between training and test.

It is worth mentioning that both vanilla and DWCCA are single models, trained on mono single channel spectrograms and no ensemble or multi-channel features were used in these experiments. In other words, a single VGG model achieves comparable performances to an ensemble of multi-channel Resnets.

We also provide class-wise f-measures on the unseen test set for both datasets in Table~\ref{tab:class_wise}.
While on the dataset without distribution shift, the average f1 stays the same by adding DWCCA in both calibrated and non calibrated models, we can observe that there is a boost of \emph{13 percentage points} on the "train" class which was the class with the lowest f1 (both calibrated and non calibrated).
It seems that DWCCA does not have a significant impact on classes with high f1: "office" and "beach" which stay the highest correctly predicted classes and do not face significant changes by DWCCA.  

On the dataset with distribution shift, we can see a significant improvement of 4 and 7 percentage points on average f1 for non-calibrated and calibrated models, respectively.
The worst class in DCASE2017 was "beach" with 32\%, which was boosted by \emph{24} and \emph{37} percentage points for non-calibrated and calibrated models, respectively.
On the other hand, the best performing class, "forest path", drops by only 2 and 3 percentage points for non-calibrated and calibrated models, respectively.

From the experimental results, we may thus conclude that overall, reducing the within-class covariance of representations using DWCCA results in more robust performance and, 
in case of a large gap between training and test, DWCCA can improve the generalisation. Additionally, the networks tend to reach a more uniform performance across various classes by improving the performance on the worst classes while not significantly degrading the best performing classes.

\begin{table}[t!]
\centering
\caption{Audio scene classification accuracy of different methods on DCASE2016 and DCASE2017 datasets. $\ast$: Single Model and single channel. $\dagger$: Ensemble of various models. $\ddagger$: Multi-channel augmentation.}
\label{tab:full_results}
\begin{tabular}{l|c|c|cccc|c|}
                           &                           &            & fold 1      & fold 2   & fold 3    & fold 4   & test     \\\hline
\multirow{5}{*}{\rotatebox[origin=r]{90}{DCASE'16}} & \multirow{2}{*}{no calib} & vanilla $\ast$   & 75.39       & 66.80   & 78.52  & 75       & 84.36 \\
                           &                           & dwcca $\ast$      & 82.42      & 75.78 & 87.89   & 84.37   & 83.60  \\\cline{2-8}
                           & \multirow{2}{*}{calib}    & vanilla $\ast$   & 76.55      & 67.59   & 78.27    & 76.21  & 85.64   \\
                           &                           & dwcca $\ast$      & 86.55      & 77.24  & 83.79   & 86.90  & 85.90  \\\cline{2-8}
                           &                           & Best CNN $\dagger$\cite{Marchi2016}     & 84.40        & 78.20     & 82.70      & 80.80     & 86.40  \\\hline
\multirow{5}{*}{\rotatebox[origin=c]{90}{DCASE'17}} & \multirow{2}{*}{no calib} & vanilla $\ast$   & 82.03    & 73.35    & 75.95    & 82.55 & 62.96 \\
                           &                           & dwcca $\ast$     & 84.03 & 81.08  & 79.43  & 87.41 & 66.23 \\\cline{2-8}
                           & \multirow{2}{*}{calib}    & vanilla $\ast$   & 82.73      & 73.93 & 76.50   & 84.61  & 62.47  \\
                           &                           & dwcca $\ast$     & 87.18     & 84.61 & 81.28 & 87.26  & 68.70  \\\cline{2-8}
                           &                           & Best CNN $\dagger\ddagger$~\cite{Zhao2017}     & 86.00     & 87.80  & 82.60   & 86.00  & 70.00      \\\hline
\end{tabular}
\end{table}

\section{Conclusion}
\label{sec:conc}
In this paper, we presented the DWCCA layer, a DNN compatible version of the classic WCCN which is used to normalize the within-class covariance of the DNN's representation and improve the performance of CNNs on data-points with shifted distributions.
Using DWCCA, we improved the performance of the VGG network by around 6 percentage point when the test datapoints were from a shifted distribution.
We analysed the embedding's generalisation by reporting KNN classification accuracies and showed that DWCCA also improves the generalisation of DNN representations both for with and without distribution mismatch.
We also showed that large within-class covariance of representations can be a sign for bad generalisation and showed that DWCCA can significantly reduce WCC and improve generalisation of the representations.

\begin{table}[t!]
\centering
\caption{The class-wise f1-score comparing the effect of adding DWCCA to Vanilla VGG on DCASE-2016 and DCASE-2017 unseen test sets. Worst-performing classes in vanilla (which benefit most from DWCCA) are shown in bold}
\label{tab:class_wise}
\begin{tabular}{l|cc|cc|cc|cc|}
 & \multicolumn{4}{|c|}{DCASE2016}                            & \multicolumn{4}{c|}{DCASE2017}                            \\\cline{2-9}
                  & \multicolumn{2}{c|}{no calib} & \multicolumn{2}{c|}{calib} & \multicolumn{2}{c|}{no calib} & \multicolumn{2}{c|}{calib} \\\cline{2-9}
                  & vanilla        & dwcca       & vanilla      & dwcca      & vanilla        & dwcca       & vanilla      & dwcca      \\\hline
beach             & 0.96     & 0.96  & 0.98    & 0.96      & \textbf{ 0.32}     & \textbf{ 0.56}  &  \textbf{0.32}    &  \textbf{0.69}  \\
bus               & 0.88     & 0.84  & 0.88    & 0.9       & 0.67     & 0.66  & 0.68    & 0.67  \\
cafe/restaurant   & 0.62     & 0.54  & 0.68    & 0.62      & 0.69     & 0.77  & 0.64    & 0.76  \\
car               & 0.94     & 0.96  & 0.96    & 0.96      & 0.83     & 0.84  & 0.84    & 0.84  \\
city\_center      & 0.94     & 0.86  & 0.93    & 0.81      & 0.73     & 0.66  & 0.72    & 0.67  \\
forest\_path      & 0.96     & 0.98  & 0.98    & 0.98      & 0.89     & 0.87  & 0.89    & 0.86  \\
grocery\_store    & 0.81     & 0.79  & 0.84    & 0.84      & 0.58     & 0.69  & 0.59    & 0.73  \\
home              & 0.93     & 0.92  & 0.91    & 0.91      & 0.69     & 0.7   & 0.67    & 0.7   \\
library           & 0.63     & 0.56  & 0.62    & 0.64      & \textbf{0.37}     &  \textbf{0.54} &  \textbf{0.36}    &  \textbf{0.61}  \\
metro\_station    & 0.75     & 0.79  & 0.81    & 0.85      & 0.47     & 0.51  & 0.45    & 0.57  \\
office            & 1        & 1     & 0.98    & 1         & 0.77     & 0.72  & 0.76    & 0.72  \\
park              & 0.77     & 0.78  & 0.77    & 0.77      & 0.39     & 0.36  & 0.41    & 0.46  \\
residential\_area & 0.79     & 0.78  & 0.82    & 0.77      & 0.49     & 0.48  & 0.5     & 0.53  \\
train             &  \textbf{0.60}      &  \textbf{0.73}  & \textbf{ 0.62 }   &  \textbf{0.87}      & 0.78     & 0.86  & 0.76    & 0.82  \\
tram              & 0.93     & 0.91  & 0.93    & 0.94      & 0.65     & 0.63  & 0.64    & 0.59  \\\hline
avg               & 0.83     & 0.83  & 0.85    & 0.85      & 0.62     & 0.66  & 0.61    & 0.68  \\
\end{tabular}
\end{table}

\section{Acknowledgment}
\label{sec:ack}
This work was supported by the Austrian Ministry for Transport, Innovation \& Technology, the Ministry of Science, Research \& Economy, and the Province of Upper Austria in the frame of the COMET center SCCH.
The authors would like to thank Hasan Bahari of KU Leuven for feedback and fruitful discussions.
We also gratefully acknowledge the support of the NVIDIA Corporation with the donation of a Titan X GPU used for this research.

\vfill\pagebreak
\bibliographystyle{plain}
\bibliography{nips_2018}

\small

\end{document}